\documentclass[journal]{IEEEtran}

\usepackage{multirow}

\usepackage[pdftex]{graphicx}
\ifCLASSINFOpdf
\else
\fi

\usepackage{array}
\begin{document}
\title{Exploring Multi-Branch and High-Level Semantic Networks for Improving Pedestrian Detection}

\author{Jiale~Cao,
        Yanwei~Pang,~\IEEEmembership{Senior~Member,~IEEE},
        and~Xuelong~Li,~\IEEEmembership{Fellow,~IEEE}% <-this % stops a space
\thanks{J. Cao and Y. Pang are with the School of Electrical and Information Engineering, Tianjin University, Tianjin 300072, China. (E-mail: \{connor,pyw\}@tju.edu.cn).}% <-this % stops a space
\thanks{X. Li is with the Center for OPTical IMagery Analysis and Learning (OPTIMAL), State Key Laboratory of Transient Optics and Photonics, Xi'an Institute of Optics and Precision Mechanics, Chinese Academy of Sciences, Xi'an 710119, Shaanxi, P. R. China. E-mail: xuelong{\_}li@outlook.com}}


\markboth{}%
{Cao \MakeLowercase{\textit{et al.}}: Exploring Multi-Branch and High-Level Semantic Networks for Improving Pedestrian Detection}



\maketitle

\begin{abstract}
To better detect pedestrians of various scales, deep multi-scale methods usually detect pedestrians of different scales by different in-network layers. However, the semantic levels of features from different layers are usually inconsistent.  In this paper, we propose a multi-branch and high-level semantic network by gradually splitting a base network into multiple different branches. As a result, the different branches have the same depth and the output features of different branches have similarly high-level semantics. Due to the difference of receptive fields, the different branches are suitable to detect pedestrians of different scales. Meanwhile, the multi-branch network does not introduce additional parameters by sharing convolutional weights of different branches. To further improve detection performance, skip-layer connections among different branches are used to add context to the branch of relatively small receptive filed, and dilated convolution is incorporated into part branches to enlarge the resolutions of output feature maps.
When they are embedded into Faster RCNN architecture, the weighted scores of proposal generation network and proposal classification network are further proposed. Experiments on KITTI dataset, Caltech pedestrian dataset, and Citypersons dataset demonstrate the effectiveness of proposed method. On these pedestrian datasets, the proposed method achieves state-of-the-art detection performance. Moreover, experiments on COCO benchmark show the proposed method is also suitable for general object detection.
\end{abstract}

\begin{IEEEkeywords}
Pedestrian detection, multi-branch network, high-level semantic features, receptive field.
\end{IEEEkeywords}

% \fi
\IEEEpeerreviewmaketitle

\section{Introduction}
\IEEEPARstart{P}{edestrian} detection plays an important role in many computer vision applications (e.g., video surveillance and driving assistance). Recently, deep convolutional neural networks (i.e., CNN) based methods have greatly promoted the progress of pedestrian detection \cite{Hosang_DeepLook_CVPR_2015,Mao_HyperLearner_CVPR_2017,Zhang_CityPersions_CVPR_2017,Sermanet_PedUMFL_CVPR_2013,Tian_Ta_CVPR_2015,Ren_RRC_CVPR_2017}. To better deal with scale-variance problem of pedestrian detection, multi-scale methods based on CNN have been proposed \cite{Ross_Fast_ICCV_2015,Girshick_RCNN_CVPR_2014,Ren_Faster_NIPS_2015,Cai_MSCNN_ECCV_2016,Lin_FPN_CVPR_2017,Kong_RON_CVPR_2017}. These methods can be mainly divided into two classes: (1) image pyramid based methods and (2) feature pyramid based methods.

Image pyramid based methods \cite{Girshick_RCNN_CVPR_2014,Ross_Fast_ICCV_2015,Ren_Faster_NIPS_2015} usually re-sample the input image into different scales and then put these rescaled images into the trained network, respectively. Thus, these methods are time-consuming. Instead of resampling the input image, feature pyramid based methods \cite{Cai_MSCNN_ECCV_2016,Liu_SSD_ECCV_2016,Kong_RON_CVPR_2017,Lin_FPN_CVPR_2017,Fu_DSSD_ARXIV_2017,Lin_Focal_ICCV_2017} use the convolutional layers of different spatial resolutions to detect pedestrians of different scales. Compared to image pyramid based methods, feature pyramid based methods make better use of features from different convolutional layers and have faster detection speed. Thus, feature pyramid based methods become more popular.

\begin{figure}[t]
\begin{center}
   \includegraphics[width=1.0\linewidth]{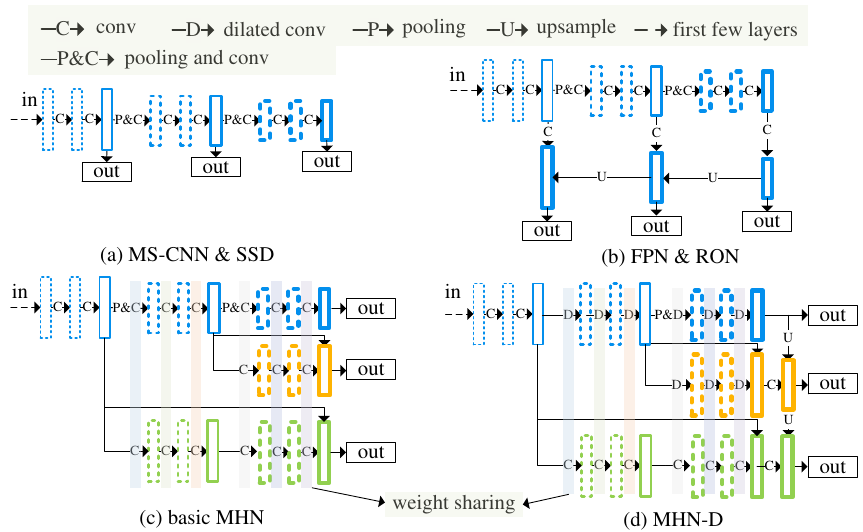}
\end{center}
   \caption{Architectures of some feature pyramid based methods. The rectangle represents feature map, and the line thickness represents the semantic level of feature map. (a) MSCNN \cite{Cai_MSCNN_ECCV_2016} and SSD \cite{Liu_SSD_ECCV_2016} use the feature maps from different in-network layers to build feature pyramid. The semantic levels of different maps are inconsistent. (b) Based on top-down structure, FPN \cite{Lin_FPN_CVPR_2017} and RON \cite{Kong_RON_CVPR_2017} build feature pyramid where all the output maps have relatively high-level semantic features. (c) The proposed basic MHN (MHN-noskip) gradually splits the base network into multiple different branches which have same depth.  Three branches can share convolutional parameters in the same column which is represented by a light color rectangle. (d) Based on basic MHN, MHN-D uses skip-layer connections among different branches and dilated convolution to further improve detection performance.}
\label{fig01}
\end{figure}

Fig. \ref{fig01} shows the architectures of some typical feature pyramid based methods. MSCNN \cite{Cai_MSCNN_ECCV_2016} in Fig. \ref{fig01}(a) uses the in-network layers of different spatial resolutions to generate the candidate proposals of different scales. SSD \cite{Liu_SSD_ECCV_2016} adopts the similar strategy for object detection. However, the maps from different layers have features of different semantic levels.  To solve this problem, FPN \cite{Lin_FPN_CVPR_2017} and RON \cite{Kong_RON_CVPR_2017} in Fig. \ref{fig01}(b) adopt top-down structure to combine the low-resolution but high-level semantic map with the high-resolution but low-level semantic map. Compared to MSCNN \cite{Cai_MSCNN_ECCV_2016}, these methods generate a feature pyramid where all the maps have relatively high-level semantic features. Despite the success, we argue that the features of combined map are not deep enough.

To further improve detection performance, we propose a multi-branch and high-level semantic convolutional neural network (called basic MHN in Fig. \ref{fig01}(c)). The base network in basic MHN (i.e., MHN-noskip) is gradually split into multiple branches before different pooling layers. Thus, all branches have the same depth, and the output maps from different branches have similarly high-level semantic features. Meanwhile, different branches have different spatial resolutions and different receptive fields, which are suitable to detect pedestrians of different scales. Moreover, MHN is generated by skip-layer connections where the high-resolution but small receptive field map is enhanced with context information by the low-resolution but large receptive field map. Because dilated convolution can retain more spatial information, it is also incorporated into part branches of MHN (i.e., MHN-D). Finally, the weighted scores of proposal generation network (RPN) and proposal classification network (Fast RCNN) are proposed when they are embedded into Faster RCNN architecture \cite{Ren_Faster_NIPS_2015}. The contributions and merits of this paper can be summarized as follows. 

(1) A multi-branch and high-level semantic convolutional neural network (MHN-noskip) is proposed for multi-scale pedestrian detection. All branches have same depth, and the output maps of different branches have similarly high-level semantic features.  

(2) Context information is introduced in MHN by skip-layer connection between the high-resolution but small receptive field map and the low-resolution but large receptive field map. Meanwhile, based on dilated convolution, the output maps in MHN-D can retain more spatial information.

(3) When MHN and MHN-D are embedded into Faster RCNN architecture, the weighted scores of RPN and Fast RCNN are proposed for pedestrian detection, which is very simple and effective.

(4) Experiments are conducted on three public pedestrian datasets (i.e., Caltech \cite{Dollar_PD_PAMI_2012}, KITTI \cite{Geiger_KITTI_CVPR_2012}, and Citypersons \cite{Zhang_CityPersions_CVPR_2017}), which demonstrate that the proposed methods have better performance than other feature pyramid methods (e.g., MSCNN \cite{Cai_MSCNN_ECCV_2016} and FPN \cite{Lin_FPN_CVPR_2017}). Moreover, experiments on very challenging COCO benchmark \cite{Lin_COCO_arXiv_2014} demonstrate that the proposed method can be successfully applied to general object detection.

The rest of this paper is organized as follows. Sec. \ref{RelatedWork} will review some related works of pedestrian detection. Sec. \ref{OurMethod} introduces our proposed method. Experimental results will be shown in Sec. \ref{Experiments}. Sec. \ref{Conclusion} concludes this paper.

\section{Related work}
\label{RelatedWork}
We firstly summarize the progress of pedestrian detection and then review deep multi-scale pedestrian detection.

\subsection{The progress of pedestrian detection}
Pedestrian detection has achieved great success in the last decade \cite{Benenson_TenYears_ECCV_2014}. It can be mainly divided into three distinct classes: (1) hand-crafted features based methods, (2) CNN features based methods, (3) the pure end-to-end methods. 

At the first few years, handcrafted channel features based method is main stream. It firstly converts the color image into ten image channels (i.e., HOG+LUV), then extracts local features (e.g., \cite{Dollar_ICF_BMVC_2009,Nam_LDCF_NIPS_2014,Zhang_HF_CVPR_2016,Zhang_FCF_CVPR_2015,Zhou_FAPD_ACM_2017}) or non-local features (e.g., \cite{Cao_NNNF_CVPR_2016}), and finally trains the cascade AdaBoost classifier. To further accelerate detection speed, Doll{\'a}r \textit{et al.} \cite{Dollar_ACF_PAMI_2014} proposed to only calculate some channels of sparse scales and approximate other channels by nearby channels. Benenson \textit{et al.} \cite{Benenson_100_CVPR_2012} proposed to train multiple detectors to detect pedestrians of different scales without image rescaling.

With the success of CNN on object detection \cite{Bell_IONet_arXiv_2015,Girshick_RCNN_CVPR_2014,Shrivastava_OHEM_CVPR_2016,Dai_RFCN_NIPS_2016,Hao_SAFD_CVPR_2017} and image classification \cite{Krizhevsky_AlexNet_NIPS_2012,Huang_DenseNet_CVPR_2017,Hu_SENet_arxiv_2017,Cheng_QCNN_TNNLS_2018,Jiang_CSN_TNNLS_2017,Marquez_DCL_TNNLS_2018}, researchers explored CNN features instead of handcrafted features for improving pedestrian detection \cite{Yang_CCF_ICCV_2015,Cai_CompACT_ICCV_2015,Tian_DeepParts_ICCV_2015,Tian_Ta_CVPR_2015,Cao_MCF_TIP_2017}. To improve representative ability of the features, Yang \textit{et al.} \cite{Yang_CCF_ICCV_2015} proposed to replace the handcrafted channel features by convolutional channel features (CCF). To have a good trade-off between accuracy and complexity, Cai \textit{et al.} \cite{Cai_CompACT_ICCV_2015} proposed CompACT cascades, which learn the efficient and handcrafted features at the former stage and the high-complexity CNN features at the later stage. Tian \textit{et al.} \cite{Tian_DeepParts_ICCV_2015} trained multiple deep part detectors and used a linear SVM to weight the detector scores.

With the success of Faster RCNN architecture on general object detection \cite{Ren_Faster_NIPS_2015}, the pure end-to-end methods have been also proposed for specific pedestrian detection. Zhang \textit{et al.} \cite{Zhang_CityPersions_CVPR_2017} made some modifications of vanilla Faster RCNN for improving pedestrian detection. Wang \textit{et al.} \cite{Wang_AFAST_CVPR_2017} utilized an adversarial sub-network to generate the occlusion and deformation positive examples in the training stage. Mao \textit{et al.} \cite{Mao_HyperLearner_CVPR_2017} proposed Hyperlearner, which joins two different tasks (i.e., semantic segmentation and pedestrian detection) into a multi-task framework. 

\subsection{Deep multi-scale pedestrian detection}
How to detect pedestrians of various scales is a challenging problem. At first, Faster RCNN \cite{Ren_Faster_NIPS_2015} does not achieve state-of-the-art performance on pedestrian detection. The main reason can be summarized as follows: (1) The spatial resolution of feature map from output layer  is low. It results that spatial information of small-scale pedestrian is heavily lost. (2) Faster RCNN generates candidate proposals of different scales based on the same feature map from last convolutional layer. As a result, the receptive field of this map can not match pedestrians of all scales very well. 

To solve above problems, many multi-scale methods have been proposed \cite{Cai_MSCNN_ECCV_2016,Lin_FPN_CVPR_2017,Liu_SSD_ECCV_2016,Kong_RON_CVPR_2017,Li_SA_arXiv_2015,Yang_SDP_CVPR_2016,Zagoruyko_MPOD_BMVC_2016}. SAF RCNN \cite{Li_SA_arXiv_2015} uses two similar sub-networks to classify small-scale pedestrians and large-scale pedestrians, respectively. SDP \cite{Yang_SDP_CVPR_2016} feds the proposals into different ROI pooling layers according to the scales of proposals. MultiPath \cite{Zagoruyko_MPOD_BMVC_2016} uses skip-layer connections and foveal regions to exploit multi-scale information. MSCNN \cite{Cai_MSCNN_ECCV_2016} generates the candidate proposals of different scales by the feature maps from different convolutional layers and then attaches detection sub-network to the relatively large feature map. To improve the semantic levels of output feature maps, FPN \cite{Lin_FPN_CVPR_2017} uses top-down structure to combine the high-level semantic but low-resolution feature map with the high-resolution but low-level semantic feature map. However, because the features of one input map are relatively low-level, the features of combined map are still not deep enough.

Meanwhile, dilated convolution (also called atrous convolution) can adjust receptive field without reducing spatial resolution of feature map. Namely, it can retain much more spatial information. Thus, dilated convolution becomes popular for object detection \cite{Yu_DRN_CVPR_2017,Zhang_RPNBF_ECCV_2016} and semantic segmentation \cite{Chen_DeepLab_PAMI_2017,Chen_Rethinking_arXiv_2017,Yu_MSCA_ICLR_2016}. Yu \textit{et al.} \cite{Yu_DRN_CVPR_2017} proposed dilated residual networks. It removes last few pooling layers and uses cascaded dilated convolutions to replace standard convolutions. Thus, it can retain more information for detection and location. Despite the success, the large receptive field cannot match pedestrians of different scales very well, especially small-scale pedestrians.

\section{The proposed methods}
\label{OurMethod}
In this section, we firstly introduce the proposed multi-branch and high-level semantic convolutional neural networks (i.e., MHN-noskip, MHN, MHN-D), and then give the illustration that how to embed the proposed multi-branch networks into the famous Faster RCNN architecture with the weighted scores of proposal generation network (i.e., RPN) and proposal classification network (i.e., Fast RCNN).

\subsection{Multi-branch and high-level semantic network}
This paper aims to generate multi-branch and high-level semantic convolutional networks for pedestrian detection. The proposed basic multi-branch  and high-level semantic network are firstly given. Based on the basic MHN (i.e., MHN-noskip), skip-layer connections and dilated convolution are further used for improving detection performance. 

\begin{figure}[t]
\begin{center}
   \includegraphics[width=1.0\linewidth]{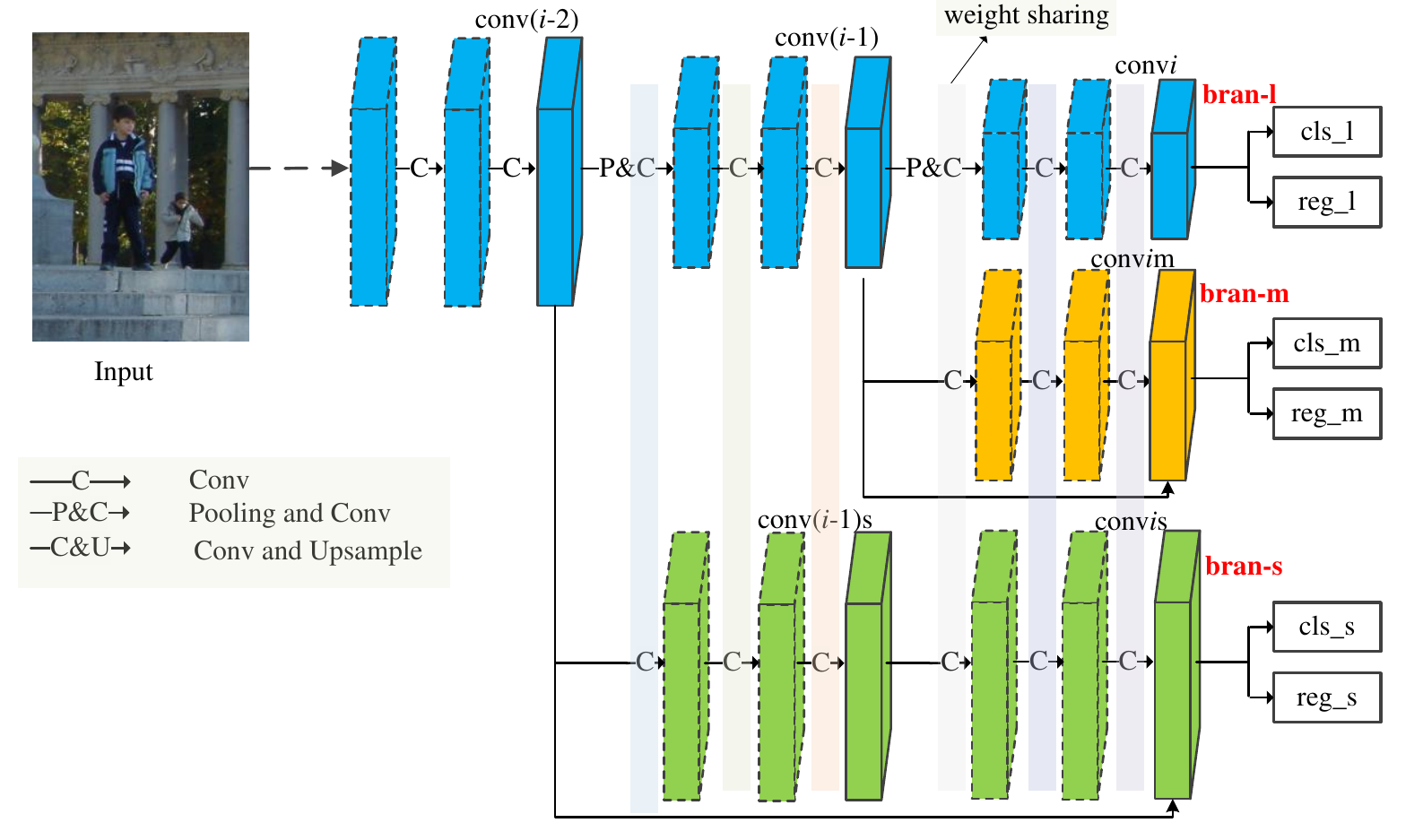}
\end{center}
   \caption{Architecture of basic MHN (i.e., MHN-noskip). The base network is gradually split into multiple different branches (i.e., bran-s, bran-m, and bran-l). In each branch, there are two sibling outputs: classification (i.e., cls) and box regression (i.e., reg).  Three branches can share the weights of convolutional layers in the same column which is represented by a light color rectangle.}
\label{fig02}
\end{figure}

\textbf{Basic MHN (called MHN-noskip)} Fig. \ref{fig02} shows the architecture of MHN-noskip which consists of three different branches (i.e., bran-s, bran-m, and bran-l). The three branches share the first few convolutional blocks (i.e., conv1, conv2, ...,conv($i$-2)). After conv($i$-2), the network is split into two different branches. One branch of bran-s goes through conv($i$-1)s and conv$i$s without pooling layer, and then combines the output feature maps of conv$i$s and conv($i$-2) by element-wise addition. Thus, the spatial resolutions of maps in bran-s are not changed and relatively largest. Another branch goes through a pooling layer and conv($i$-1). After conv($i$-1), this branch is further split into two different branches (i.e., bran-m and bran-l). The branch of bran-m further goes through conv$i$m without pooling layer and combines the output feature maps of conv$i$m and conv($i$-1), while the branch of bran-l goes through a pooling layer and conv$i$.

On the one hand, the spatial resolutions of output feature maps in bran-s, bran-m, bran-l gradually decrease, while the receptive fields of output feature maps in bran-s, bran-m, bran-l gradually increase. Thus, bran-s with the largest spatial resolution and smallest receptive field is suitable for small-scale pedestrian detection, bran-m is used for medium-scale pedestrian detection, and bran-l with the smallest spatial resolution and largest receptive field is for large-scale pedestrian detection. On the other hand, the three different branches (i.e., bran-s, bran-m, and bran-l) undergoes same number of convolutional layers. Thus, the different branches have same depth, and the output maps of different branches have similarly high-level semantic features. Please note that bran-s, bran-m, and bran-l can share the weights of convolutional layers in the same column. In Fig. \ref{fig02}, the same column is represented by a light color rectangle. 

\begin{figure}[t]
\begin{center}
   \includegraphics[width=1.0\linewidth]{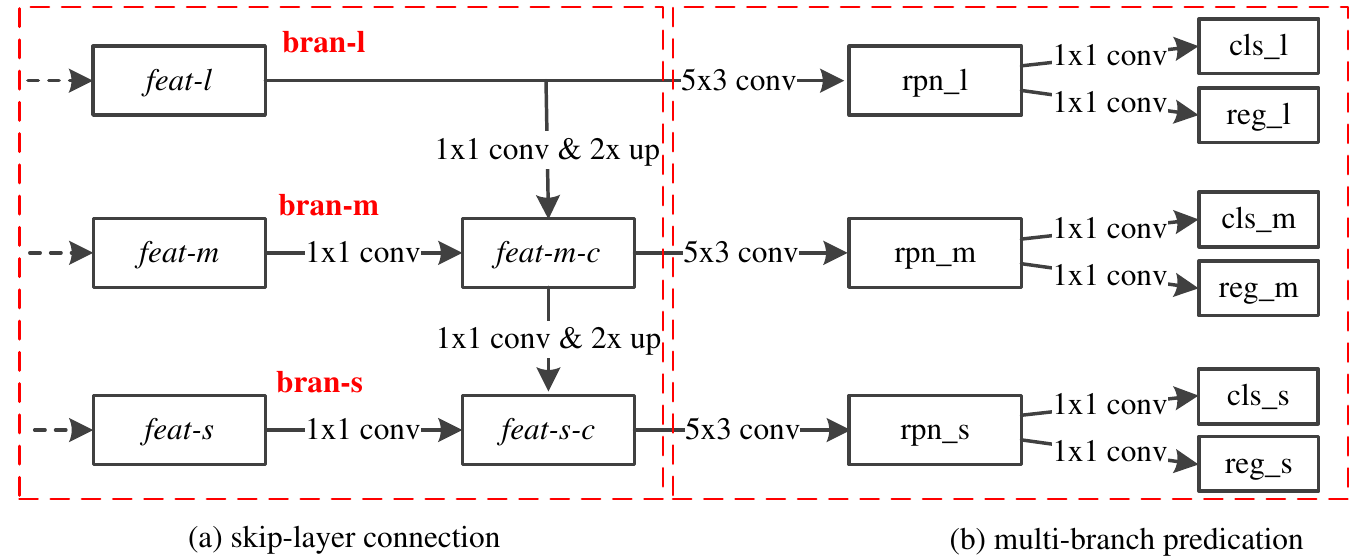}
\end{center}
   \caption{(a) Skip-layer connections among different branches (i.e., MHN). The feature map \textit{feat-m-c} is the combination of the map \textit{feat-l} from bran-l and the map \textit{feat-m} from bran-m, then the feature map \textit{feat-s-c} is the combination of the map \textit{feat-l} from bran-l, the map \textit{feat-m} from bran-m, and the map \textit{feat-s} from bran-s. (b) Multi-branch predications of bran-l, bran-m, and bran-s.}
\label{fig03}
\end{figure}

\textbf{Skip-layer connections among different branches (called MHN)} Context information have been demonstrated to be useful for small-scale object detection \cite{Chen_Rethinking_arXiv_2017,Zhao_PSPNet_CVPR_2017}. To introduce the context information for small-scale pedestrian detection, skip-layer connections are used to combine the output feature maps of different branches. Fig. \ref{fig03}(a) shows skip-layer connections among different branches. \textit{feat-l}, \textit{feat-m}, \textit{feat-s} represent the output maps of bran-l, bran-m, bran-s, respectively. It can be seen that the feature map \textit{feat-m-c} is fused by the map \textit{feat-l} and the map \textit{feat-m}. Specifically, the map \textit{feat-l} undergoes a $1\times1$ convolutional layer and a deconvolutional layer which upsamples the input map twice by bilinear interpolation; the map \textit{feat-m} undergoes a $1\times1$ convolutional layer; after that, the map \textit{feat-m-c} is generated by element-wise addition of the output maps of above two steps. In the similar fashion, the map \textit{feat-s-c} is generated by fusing the map \textit{feat-s} and the map \textit{feat-m-c}. 

Because the output feature map of bran-l (i.e., \textit{feat-l}) has larger receptive field than the output feature map of bran-m (i.e., \textit{feat-m}), the combination of \textit{feat-l} and \textit{feat-m} (i.e., \textit{feat-m-c}) can add context information to the output feature map of bran-m. Namely, context information can be added to bran-m by bran-l. Similarly, context information can be cumulatively added to bran-s by bran-m and bran-l. Meanwhile, because the input maps of the combination (i.e., \textit{feat-s}, \textit{feat-m}, \textit{feat-l}) have high-level semantic features, the combined maps (i.e., \textit{feat-m-c}, \textit{feat-s-c}) have much deeper features.

Compared to MSCNN \cite{Cai_MSCNN_ECCV_2016}, the semantics of output maps in MHN are consistent and high-level. Compared to FPN \cite{Lin_FPN_CVPR_2017}, the input maps of combination in MHN all have high-level semantic features. As a result, the combined maps in MHN have much deeper features. In summarization, MHN has the following advantages: (1) Multi-branch structure is suitable for multi-scale pedestrian detection. Different branches have different receptive fields and spatial resolutions, which can better match pedestrians of different scales; (2) The skip-layer connections among different branches can enhance context information for small-scale pedestrian detection. Meanwhile, the fused features are much deeper. 

\textbf{MHN with dilated convolution (called MHN-D)} Dilated convolution can enlarge spatial resolution of output feature map without reducing its receptive field. Recently, dilated convolution is used for object detection. For example, Yu \textit{et al.} \cite{Yu_DRN_CVPR_2017} removed some pooling layers and replaced the following standard convolutions by dilated convolutions. Despite the success, the large receptive field of output feature map does not match small-scale pedestrians very well. Thus, in our proposed architecture (called MHN-D), dilated convolution is only used in part branches of the multi-branch network.

\begin{figure}[t]
\begin{center}
   \includegraphics[width=1.0\linewidth]{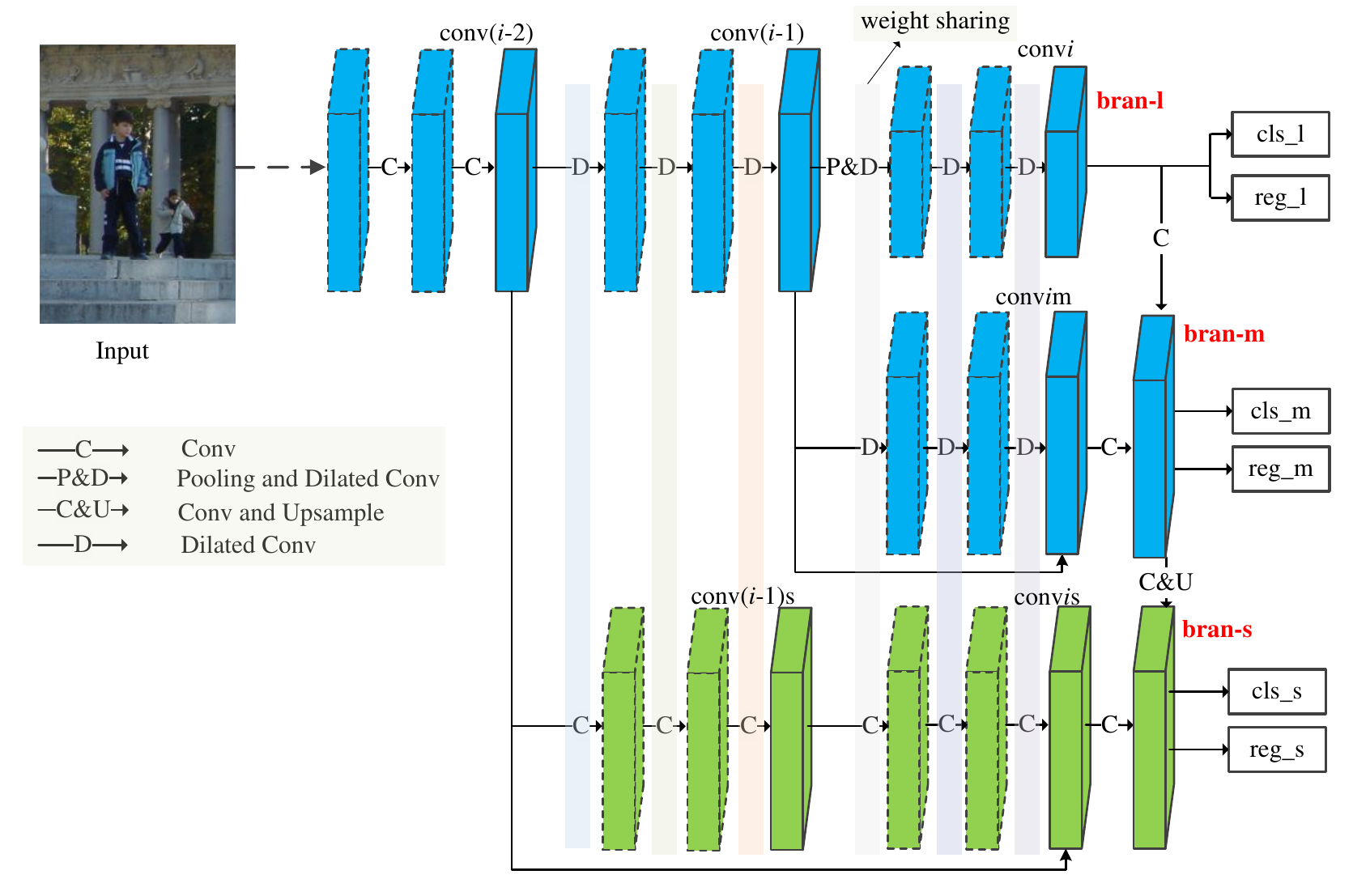}
\end{center}
   \caption{Architecture of MHN-D. It removes the pooling layer of pool($i$-2) and incorporates dilated convolutions into MHN to enlarge the spatial resolutions of bran-l and bran-m without reducing receptive fields.}
\label{fig04}
\end{figure}

Fig. \ref{fig04} shows the architecture of MHN-D.  The pooling layer (i.e., pool($i$-2)) after conv($i$-2) is removed, and the standard convolutions in conv($i$-1), conv$i$, and conv$i$m are replaced by dilated convolution. As a result, the spatial resolutions of output maps in bran-m and bran-l of MHN-D are twice larger. More spatial information can be retained for object location and thus improve detection performance.

\textbf{Multi-branch predictions} As described above, three different branches (i.e., bran-s, bran-m, bran-l) detect pedestrians of different scales, respectively. Fig. \ref{fig03}(b) shows the head networks for multi-branch predictions. They are attached to the output maps of each branch. The structures of these head networks are same, which consist of a $5\times3$ convolutional layer and two sibling $1\times1$ convolutional layers for classification and box regression. Please note that, instead of a $3\times3$ convolution, the $5\times3$ convolution is used to better match the shape of pedestrians.

Assuming that the pedestrian scales belong to [$s_{min}$, $s_{max}$], pedestrians are then equally divided into $N$ bins in the logarithmic space. Then, the scales of $N$ anchors are set as the centers of $N$ bins, respectively. Namely, the scale of $n$-th anchor (i.e., $s_n$) can be written as follows.
\begin{equation}
s_n=s_{min} \times \{\frac{s_{max}}{s_{min}}\}^{(n-0.5)/N}, n=1,2,...,N.
\end{equation}
On KITTI dataset and Caltech pedestrian dataset, the number of anchors (i.e., $N$) is set as 9. Thus, the three anchors of the smallest scales (i.e, $s1$, $s2$, and $s3$) belong to bran-s, the three anchors of medium scales (i.e., $s4$, $s5$, and $s6$) belong to bran-m, and the three anchors of largest scales (i.e., $s7$, $s8$, and $s9$) belong to bran-l. Because the scales of pedestrians on Citypersons range in a much larger space, the number of anchors is set as 12. Thus, the four anchors of the smallest scales (i.e, $s1$, $s2$, $s3$, and $s4$) belong to bran-s, the four anchors of medium scales (i.e., $s5$, $s6$, $s7$, and $s8$) belong to bran-m, and the four anchors of largest scales (i.e., $s9$, $s10$, $s11$, and $s12$) belong to bran-l. Different from general object detection, the aspect ratio of anchors is only set as a fixed value which is calculated by the average aspect ratios of pedestrians on the dataset. Specifically, the aspect ratios are set as $1/0.41$ for Caltech pedestrian dataset and Citypersons dataset, and the aspect ratio is set as $1/0.36$ for KITTI dataset.

\subsection{Weighted classification scores when embedded into Faster RCNN architecture}

\begin{figure}[t]
\begin{center}
   \includegraphics[width=1.0\linewidth]{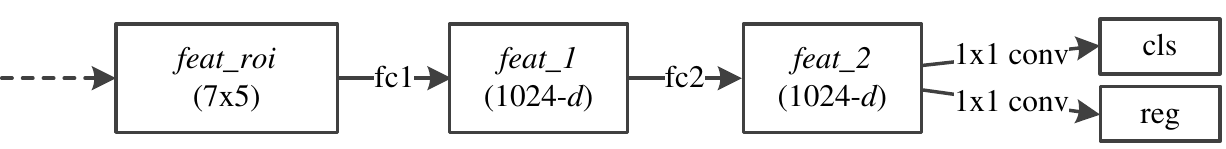}
\end{center}
   \caption{Architecture of Fast RCNN head network for pedestrian detection.}
\label{fig05}
\end{figure}

To further improve detection performance, MHN and MHN-D are embedded into Faster RCNN architecture \cite{Ren_Faster_NIPS_2015}. Faster RCNN consists of two parts: proposal generation network (RPN) and proposal classification network (Fast RCNN). RPN subnetwork is replaced by MHN or MHN-D. They are used to extract pedestrian proposals, and then these proposals are classified by Fast RCNN. Fig. \ref{fig05} shows the Fast RCNN head network for pedestrian detection. Similar to \cite{Cai_MSCNN_ECCV_2016}, the Fast RCNN head network is attached to the high-resolution feature map of multi-branch network for pedestrian detection.  

In Faster RCNN architecture \cite{Ren_Faster_NIPS_2015}, Fast RCNN ignores the classification scores of RPN when classifying these proposals. Recently, some successful bootstrap techniques (e.g., OHEM \cite{Shrivastava_OHEM_CVPR_2016} and FL \cite{Lin_Focal_ICCV_2017}) are used to generate hard proposals in the training stage. Despite the success, they still do not make full use of RPN scores.  In this paper, it is found that RPN scores can be helpful for improving performance of pedestrian detection. Namely, the final classification score (i.e., $S_f$) is the weighted sum of Fast RCNN score (i.e., $S_{rcnn}$) and MHN score (i.e., $S_{mhn}$) as follows: 
\begin{equation}
S_f=S_{rcnn}+\lambda \times S_{mhn},
\end{equation}
where $\lambda$ balances two items (i.e., RCNN score and MHN score), which is set by cross-validation. 

\section{Experiments}
\label{Experiments}
\subsection{Datasets}
\label{ExperimentsDataset}
Three public pedestrian datasets (i.e., KITTI dataset \cite{Geiger_KITTI_CVPR_2012}, Caltech pedestrian dataset \cite{Dollar_PD_PAMI_2012}, and Citypersons pedestrian dataset) taken from vehicle driving are used to demonstrate the effectiveness of the proposed method and compare with some state-of-the-art methods. To demonstrate suitability for general object detection, experiments on COCO benchmark are further conducted. 

KITTI dataset \cite{Geiger_KITTI_CVPR_2012} is a challenging computer vision benchmark, which is used to develop autonomous driving. On KITTI dataset, pedestrian detection is one sub-task of object detection. It consists of 7481 training images and 7518 test images. To enlarge the number of images on training sets, data augmentation is used. The original image is randomly rescaled by a factor of $a\in [0.8,1.2]$ nine times.  Then, the rescaled images are cropped or padded to the same size of the original image. Average precision ($AP$) is used as the evaluation criterion.

Caltech pedestrian dataset \cite{Dollar_PD_PAMI_2012} consists of 11 videos, where the first 6 videos are training sets and the last 5 videos are test sets. To enlarge the number of training images, the training images are captured by sampling one frame per 3 frames. The test images are captured following the unified standard, which samples one frame per 30 frames. In \cite{Dollar_PD_PAMI_2012}, it states that detecting pedestrians over 30 pixels are necessary for autonomous driving. Thus, pedestrians over 30 pixels instead of 50 pixels are used for evaluation. Log-averaged miss rates ($MR$) over FPPI=[0.01,1] are used as the evaluation criterion.

Citypersons pedestrian dataset \cite{Zhang_CityPersions_CVPR_2017} is built on Cityscapes benchmark \cite{Cordts_Cityscapes_CVPR_2016}. With 5000 fine annotations subset on Cityscapes, high-quality annotations of pedestrians are created by Citypersons. Specifically, Citypersons contains three subsets (i.e.,  train, validation, and test sets). Similar to Caltech pedestrian dataset \cite{Dollar_PD_PAMI_2012}, pedestrians over 30 pixels are used for evaluation and log-averaged miss rates ($MR$) over FPPI=[0.01,1] are used as the evaluation criterion.

COCO benchmark \cite{Lin_COCO_arXiv_2014} is a large-scale and challenging  object detection, segmentation, and captioning dataset, which has 80 object categories. The images are split into 80k training images and 40k validation images. Usually, 80k training images and a 35k subset of validation images (i.e., \texttt{trainval35k}) are used for training. A 5k subset of validation images (i.e., \texttt{minival}) are used for performance evaluation.

\subsection{Some middle experimental results}
\label{ExperimentsMiddle}
In this subsection, multi-branch and high-level semantic networks (i.e., MHN-noskip, MHN, MHN-D) are constructed based on VGG16 \cite{Simonyan_VGG_arxiv_2014} and the new added pool5 and conv6. The weights of conv1 to conv5 in MHN are initialized based on pre-trained model. The convolutional layers of conv6 in MHN are initialized from Gaussian distribution with a standard deviation of 0.01. The convolutional weights of conv5s are copied from conv5. The convolutional weights of conv6s and conv6m are both copied from conv6. 

Two related and representative feature pyramid methods (i.e., MSCNN \cite{Cai_MSCNN_ECCV_2016} and FPN \cite{Lin_FPN_CVPR_2017}) are compared to our proposed methods (i.e., MHN-noskip, MHN, and MHN-D) on KITTI dataset, Caltech dataset, and Citypersons dataset to demonstrate the effectiveness of proposed methods. According to \cite{Chen_MO3D_CVPR_2016}, the training images on KITTI dataset are split into the training set and validation set. For fair comparison, all the above methods are trained on three datasets with the similar parameter settings. On Caltech and KITTI datasets, the number of total iterations is 70k, where the learning rate of first 50k iterations is 0.001 and the learning rate of last 20k iterations is 0.0001. On Citypersons dataset, the number of total iterations is 120k, where the learning rate of first 100k iterations is 0.001 and the learning rate of last 20k iterations is 0.0001. In each mini-batch, there are only one image and 256 anchors per branch.

\begin{table}[t]
\renewcommand{\arraystretch}{1.3}
\caption{Detection results of MHN-share and MHN-noshare on three pedestrian datasets. The difference between MHN-share and MHN-noshare is whether sharing weights among different branches.}
\begin{center}
\footnotesize
\begin{tabular}{l|ccc}
\hline
\multirow{2}{*}{Method}             & KITTI             & Caltech    & Citypersons \\
\cline{2-4}
                                    & $AP$ (\%)         & $MR$ (\%)   & $MR$ (\%) \\
\hline                
MHN-noshare                         & 65.99             & 41.80        & 34.13\\
MHN-share                           & 65.96             & 42.07        & 34.04\\
\hline
\end{tabular}
\end{center}
\label{tab01}
\end{table}
\begin{table}[t]
\renewcommand{\arraystretch}{1.3}
\caption{Detection results and network parameters of MSCNN, FPN, and the proposed methods (i.e., MNH-noskip, MHN, and MHN-D) on three pedestrian datasets.}
\begin{center}
\footnotesize
\begin{tabular}{l|c|ccc}
\hline
\multirow{2}{*}{Method}       & \multirow{2}{*}{\#params}      & KITTI   & Caltech & Citypersons     \\
\cline{3-5}
             &                 & $AP$ (\%)     & $MR$ (\%) & $MR$ (\%) \\
\hline
MSCNN  \cite{Cai_MSCNN_ECCV_2016}      &$\sim$24M        & 63.73        & 44.57        & 35.78\\
MHN-noskip   & $\sim$24M       & 64.57        & 42.96        & 34.84\\
\hline
$\Delta$     &  N/A              & \textbf{0.84}         & \textbf{1.59}         &  \textbf{0.94}\\
\hline
\hline
FPN   \cite{Lin_FPN_CVPR_2017}       &  $\sim$30M      & 64.95        & 42.90        & 35.20\\
MHN          &  $\sim$30M      & 65.96        & 42.07        & 34.04\\
MHN-D        &  $\sim$30M      & 66.86        & 40.61        & 32.81\\
\hline
$\Delta$     &  N/A            & \textbf{1.91}         & \textbf{2.29}         &  \textbf{2.39}\\
\hline
\end{tabular}
\end{center}
\label{tab02}
\end{table}
\begin{table*}[ht]
\renewcommand{\arraystretch}{1.3}
\caption{Average precision ($AP$) of MSCNN \cite{Cai_MSCNN_ECCV_2016}, FPN \cite{Lin_FPN_CVPR_2017} and MHN on KITTI validation set with moderate difficulty. These methods are all three output layers.}
\begin{center}
\footnotesize
\begin{tabular}{p{3cm}|p{2cm}<{\centering}|p{2cm}<{\centering}|ccc|c}
\hline
Method & Output Layer & Feature Stride & $AP_s$ & $AP_m$ & $AP_l$ & $AP$ \\
\hline
MSCNN \cite{Cai_MSCNN_ECCV_2016}& C4, C5, C6 & 8, 16, 32 & 26.38& 61.52& 84.91& 63.73\\
FPN \cite{Lin_FPN_CVPR_2017}& P4, P5, P6 & 8, 16, 32     & 29.71& 62.53& 85.32& 64.95\\
MHN & M4, M5, M6 & 8, 16, 32                             & 32.47& 62.81& 85.81& \textbf{65.96}\\
\hline
\multicolumn{6}{l}{Ablation experiment}\\
\hline
no skip & M4, M5, M6 & 8, 16, 32 & 27.69& 62.17& 84.94& 64.57\\
bran-s  & M4         & 8         & 20.10& 58.69& 82.33& 61.11\\
bran-m  & M5         & 16        & 18.50& 59.74& 87.45& 63.20\\
bran-l  & M6         & 32        & 11.63& 37.42& 84.38& 53.29\\
\hline
\end{tabular}
\end{center}
\label{tab03}
\end{table*}
\begin{table*}[ht]
\renewcommand{\arraystretch}{1.3}
\caption{Average precision ($AP$) of DCN \cite{Yu_DRN_CVPR_2017}, MHN and MHN-D on KITTI validation set with moderate difficulty. For fair comparison, DCN is a variant of \cite{Yu_DRN_CVPR_2017} based on VGG16 instead of ResNet, which removes pool4 and uses the dilated convolution in conv5.}
\begin{center}
\footnotesize
\begin{tabular}{p{3cm}|p{2cm}<{\centering}|p{2cm}<{\centering}|ccc|c}
\hline
Method & Output Layer & Feature Stride & $AP_s$ & $AP_m$ & $AP_l$ & $AP$ \\
\hline
DCN \cite{Yu_DRN_CVPR_2017}& C6         & 8        & 27.31& 61.90& 86.44& 65.23\\
MHN                        & M4, M5, M6 & 8, 16, 32& 32.47& 62.81& 85.81& 65.96\\
MHN-D                      & M4, M5, M6 & 8,  8, 16& 30.92& 63.40& 87.02& \textbf{66.86}\\
\hline
\end{tabular}
\end{center}
\label{tab04}
\end{table*}
\textbf{Weight sharing among different branches} As described in Sec. \ref{OurMethod}, different branches of MHN can share convolutional weights in the same column. Table \ref{tab01} compares detection performance of MHN-share and MHN-noshare on KITTI, Caltech, and Citypersons, where MHN-share shares convolutional weights among different branches and MHN-noshare does not share convolutional weights among different branches. It can be seen that MHN-share and MHN-noshare have similar detection performance. For example, $MR$s of MHN-noshare and MHN-share on Caltech test set are 41.80\% and 42.07\%, respectively.  Because MHN-share has fewer network parameters than MHN-noshare, MHN-share is used in the following experiments.

\textbf{Compared to MSCNN and FPN} Table \ref{tab02}. compares the proposed methods (MHN-noskip, MHN, and MHN-D) with MSCNN and FPN on KITTI, Caltech, and Citypersons datasets in detail, where network parameters and detection performance are both shown. Based on Table \ref{tab02}, it can be seen: (1) MHN-noskip is significantly better than MSCNN with almost same number of network parameters. For example, MHN-noskip outperforms MSCNN by 0.84\%, 1.67\%, and 0.94\% on KITTI, Caltech, and Citypersons datasets, respectively. Meanwhile, MNH-noskip is already comparable to FPN with fewer network parameters. (2) With almost same number of network parameters, MHN outperforms FPN by 1.01\%, 0.83\%, and 1.16\% on KITTI, Caltech, and Citypersons datasets, respectively. It means that the improvement of our proposed method is not from more network parameters. (3) MHN-D has best detection performance, which outperforms MSCNN and FPN. For example, MHN-D outperforms FPN by 1.91\%, 2.29\%, and 2.39\% on KITTI, Caltech, and Citypersons datasets, respectively.

\textbf{Ablation experiments of MHN} To further demonstrate effectiveness of MHN, ablation experiments are further conducted on KITTI validation set with moderately difficult level.  The moderate set of KITTI dataset is further divided into three subsets (i.e., \texttt{small} subset, \texttt{medium} subset, and \texttt{large} subset) according to the height of pedestrians. Specifically, the \texttt{small} subset means the pedestrians under 60 pixels and over 25 pixels, the \texttt{medium} subset means the pedestrians over 60 pixels and under 120 pixels, and the \texttt{large} subset means the pedestrians over 120 pixels. Experimental results of MSCNN, FPN, MHN on the three subsets are shown in Table \ref{tab03}. They are all three output layers.  MSCNN \cite{Cai_MSCNN_ECCV_2016} outputs pedestrian proposals on C4, C5, and C6. FPN \cite{Lin_FPN_CVPR_2017} outputs pedestrian proposals on P4, P5, and P6, and MHN outputs pedestrian proposals on M4, M5, and M6. On the one hand, $AP$ of MSCNN \cite{Cai_MSCNN_ECCV_2016}, FPN \cite{Lin_FPN_CVPR_2017}, and MHN are 63.73\%, 64.95\%, and 65.96\%, respectively. Thus, MHN outperforms MSCNN and FPN by 2.23\% and 1.01\%. Namely, MHN has better performance than MSCNN and FPN. On the other hand, on small subset, $AP$ of MSCNN \cite{Cai_MSCNN_ECCV_2016}, FPN \cite{Lin_FPN_CVPR_2017}, and MHN are 26.38\%, 29.71\%, and 32.47\%, respectively. MHN outperforms MSCNN and FPN by 6.09\% and 2.76\%. It can be seen that MHN has better detection performance on small-scale pedestrian detection.

Table \ref{tab03} also shows ablation experiments of MHN. ``no skip'' means that the skip-layer connections in MHN  are removed (i.e., MHN-noskip). ``bran-s'' means that the anchors of all scales are attached to M4 and other branches are removed. ``bran-m'' attaches the anchors of all scales to M5. ``bran-l'' attaches the anchors of all scales to M6. Without skip-layer connections, $AP$ of MHN decreases by 1.39\%. On small subset, $AP$ of ``bran-l'' is 8.47\% and 6.87\% lower than that of ``bran-s'' and ``bran-m''. Because the spatial resolutions of ``bran-s'', ``bran-m'', and ``bran-l'' decrease by a factor of 2, spatial information of ``bran-l'' is largely lost. Thus, ``bran-l'' has the worst detection performance on small-scale pedestrian detection.

\textbf{Ablation experiments of MHN-D} Recently, dilated convolution has been successfully applied to object detection \cite{Yu_DRN_CVPR_2017}, which can enlarge the receptive field of feature map without reducing its spatial resolution. For example, DRN \cite{Yu_DRN_CVPR_2017} improves ResNet \cite{He_ResNet_CVPR_2016} by dilated convolution. Following DRN \cite{Yu_DRN_CVPR_2017}, dilated convolution is embedded into original VGG16 (called DCN) and MHN (called MHN-D). For DCN and MHN-D, the standard convolutions of conv5 are replaced by dilated convolutions and the pooling layer of pool4 is removed. Table \ref{tab04} compares DCN, MHN, and MHN-D. It can be concluded as follows: (1) MHN-D has better detection performance than MHN. It means that dilated convolution is useful for further improving detection performance. (2) MHN-D has better detection performance than DCN. Though the receptive field of DCN is very large, it does not match pedestrians of various scales very well. Compared to DCN, multiple different receptive fields of MHN-D are more suitable for multi-scale pedestrian detection.

\begin{table}[t]
\renewcommand{\arraystretch}{1.3}
\caption{Average precision of some methods when embedded into Faster RCNN \cite{Ren_Faster_NIPS_2015}. ${\surd}$ means that the method is based on feature pyramid, and + means that the weighted scores are used.}
\begin{center}
\footnotesize
\begin{tabular}{p{1.5cm}|c|c||c|c|c}
\hline
Method & Pyramid & $AP$ (\%) & Method & Pyramid & $AP$ (\%) \\
\hline
MSCNN \cite{Cai_MSCNN_ECCV_2016} & ${\surd}$ & 72.01 & MHN          & ${\surd}$ & 73.30\\
FPN \cite{Lin_FPN_CVPR_2017}     & ${\surd}$ & 72.65 & MHN+         & ${\surd}$ & 73.97  \\
DCN \cite{Yu_DRN_CVPR_2017}      &           & 72.79 & MHN-D+       & ${\surd}$ & \textbf{74.53}\\
\hline
\end{tabular}
\end{center}
\label{tab05}
\end{table}
\textbf{MHN and MHN-D embedded into Faster RCNN} In the training stage, the number of total iterations is 70k. The learning rate of first 50k iterations is 0.001, and that of last 20k iterations is 0.0001. In each iteration, there are only one image, 256 RPN anchors, and 256 ROIs. 

Table \ref{tab05} shows these feature pyramid methods based on Faster RCNN pipeline. + in Table \ref{tab05} means the weighted scores of RPN and Fast RCNN are used. For fair comparison, all the methods are re-implemented on the similar parameter settings. DCN \cite{Yu_DRN_CVPR_2017} outputs the proposals on the feature map from last convolutional layer, while MSCNN \cite{Cai_MSCNN_ECCV_2016}, FPN \cite{Lin_FPN_CVPR_2017}, MHN, and MNH-D generate the pedestrian proposals on multiple feature maps from different convolutional layers. Based on Table \ref{tab05}, it can be seen that MHN, MHN+ and MHN-D+ outperform other methods (i.e., MSCNN \cite{Cai_MSCNN_ECCV_2016}, FPN \cite{Lin_FPN_CVPR_2017}, and DCN \cite{Yu_DRN_CVPR_2017}). For example, MHN outperforms MSCNN, FPN, and DCN by 1.29\%, 0.65\%, 0.51\%, respectively. MHN-D+ outperforms MSCNN, FPN, and DCN by 2.52\%, 1.88\%, 1.74\%, respectively. It means that the proposed method is useful on Faster RCNN pipeline. MHN+ outperforms MHN, which means the effectiveness of the weighted scores.

\begin{table}[t]
\renewcommand{\arraystretch}{1.3}
\caption{Average precision ($AP$) and inference time ($T$) on validation set with moderate difficulty.}
\begin{center}
\footnotesize
\begin{tabular}{l|cccc}
\hline
Method & MSCNN \cite{Cai_MSCNN_ECCV_2016} & FPN \cite{Lin_FPN_CVPR_2017} & MHN & MHN-D \\
\hline
$AP$ (\%) & 63.78 & 64.93& 65.96& \textbf{66.86}  \\
$T$ (ms) & 72& 82& 98& 113\\
\hline
\multicolumn{5}{l}{Embedded into Faster RCNN}\\
\hline
$AP$ (\%) &  72.01& 72.65& 73.97& \textbf{74.53}\\
$T$ (ms)  & 76& 87 & 104& 118\\
\hline
\end{tabular}
\end{center}
\label{tab06}
\end{table}

\textbf{Inference time} Table \ref{tab06} shows average precision ($AP$) and inference time ($T$) of different methods on the same device, where the CPU is Intel Xeon E5-2695 and the GPU is NVIDIA Quadro P6000. $AP$s of FPN and MHN are 64.93\% and 65.96\%, while inference time of FPN  and MHN are 82ms and 98ms. When FPN and MHN are embedded into Faster RCNN, $AP$s of FPN and MHN are 72.65\% and 73.97\%, and inference time of FPN and MHN are 87ms and 104ms. It means that MHN can improve detection performance with little increase of computation cost. Compared to MHN, MHN-D can further improve detection performance.

\begin{table}[t]
\renewcommand{\arraystretch}{1.3}
\caption{Average precision ($AP$) and Inference time ($T$) of some state-of-the-art methods on KITTI test set with three different difficulties (i.e., Easy, Moderate, and Hard). $AP$ and $T$ of these methods are given by KITTI website.}
\begin{center}
\footnotesize
\begin{tabular}{p{2.5cm}|ccc|c}
\hline
Method & Moderate & Easy & Hard & Time\\
\hline
CompACT-Deep \cite{Cai_CompACT_ICCV_2015} & 58.73  & 69.70  & 52.69  & 1.00\\
Mono3D  \cite{Chen_MO3D_CVPR_2016}        & 66.66  & 77.30  & 63.44  & 4.20\\
RPN+BF  \cite{Zhang_RPNBF_ECCV_2016}      & 61.29  & 75.58  & 56.08  & 0.60\\
Faster RCNN \cite{Ren_Faster_NIPS_2015}   & 65.91  & 78.35  & 61.19  & 2.00\\
3DOP \cite{Chen_3DOP_NIPS_2015}           & 67.46  & 82.50  & 65.14  & 3.00\\
SDP \cite{Yang_SDP_CVPR_2016}             & 70.20  & 79.98  & 64.84  & 0.40\\
IVA \cite{Zhu_IVA_ACCV_2016}              & 70.63  & 83.03  & 64.68  & 0.40\\
SubCNN \cite{Xiang_SubCNN_WACV_2017}      & 71.34  & 83.17  & 66.36  & 2.00\\
MSCNN \cite{Cai_MSCNN_ECCV_2016}          & 73.62  & 83.70  & 68.28  & 0.40\\
RRC \cite{Ren_RRC_CVPR_2017}              & 75.33  & 84.14  & 70.39  & 3.60\\
\hline
MHN-D                                     & 74.60  & \textbf{85.81}  & 68.94  & 0.39\\
\hline
\end{tabular}
\end{center}
\label{tab07}
\end{table}
\begin{figure}[t]
\begin{center}
   \includegraphics[width=0.8\linewidth]{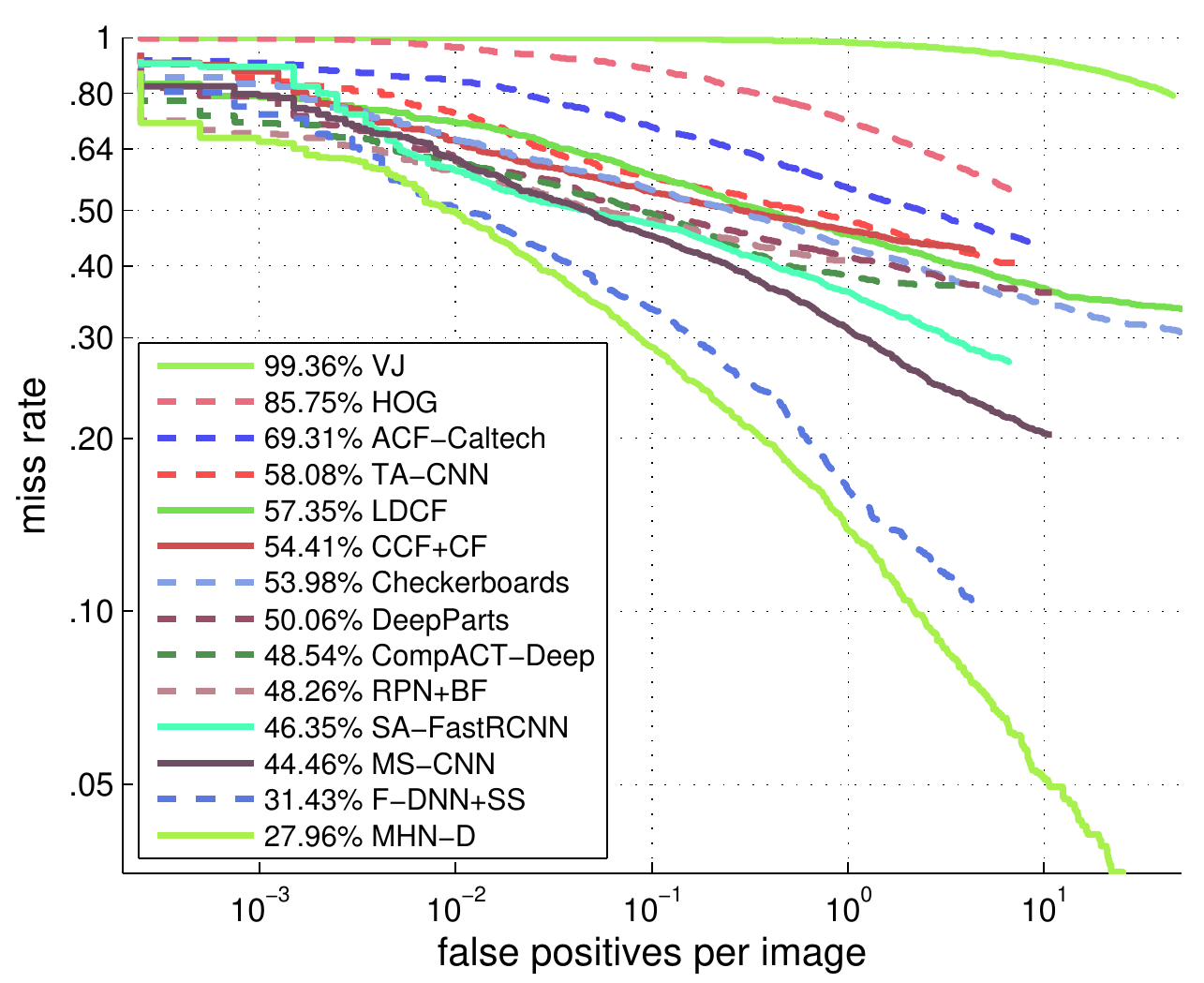}
\end{center}
   \caption{ROC on Caltech test set, where pedestrians over 30 pixels are used for evaluation. The legend represents the log-averaged miss rate over FPPI=[0.01,1].}
\label{fig06}
\end{figure}

\subsection{Comparison to some state-of-the-art methods on KITTI dataset}
In this subsection, the proposed method is compared to some state-of-the-art methods (e.g., RRC \cite{Ren_Faster_NIPS_2015}, MSCNN \cite{Cai_MSCNN_ECCV_2016}, SubCNN \cite{Xiang_SubCNN_WACV_2017}, IVA \cite{Zhu_IVA_ACCV_2016}, SDP \cite{Yang_SDP_CVPR_2016}, 3DOP \cite{Chen_3DOP_NIPS_2015}, Faster RCNN \cite{Ren_Faster_NIPS_2015}, RPN+BF \cite{Zhang_RPNBF_ECCV_2016}, Mono3D \cite{Chen_MO3D_CVPR_2016}, CompACT-Deep \cite{Cai_CompACT_ICCV_2015}) on KITTI test set \cite{Geiger_KITTI_CVPR_2012}. To achieve better detection performance, the images are twice upsampled in the training and test stages. The training parameters of MHN-D follows that described above. The subsets of three different difficulties (i.e., Easy, Moderate, and Hard) on the test images are used for evaluation. Table \ref{tab07} shows the average precision ($AP$) on the three subsets. It can be seen that the proposed MHN-D almost achieves the state-of-the-art performance. On moderate test set, it outperforms MSCNN \cite{Cai_MSCNN_ECCV_2016} and SubCNN \cite{Xiang_SubCNN_WACV_2017} by 0.98\% and 3.26\%. On easy test set, it outperforms MSCNN and SubCNN by 2.11\% and 2.64\%, respectively. Though RRC \cite{Ren_RRC_CVPR_2017} has a little better detection performance than MHN-D on moderate and hard test sets, it is about 8 times slower than MHN-D. Moreover, our proposed method can be applied to RRC.

\subsection{Comparison to some state-of-the-art methods on Caltech pedestrian dataset}
In this subsection, the proposed method is compared to some state-of-the-art methods (i.e., F-DNN+SS \cite{Du_FDNN_WACV_2017}, MSCNN \cite{Cai_MSCNN_ECCV_2016}, SA-FastRCNN \cite{Li_SA_arXiv_2015}, RPN+BF \cite{Zhang_RPNBF_ECCV_2016}, CompACT-Deep \cite{Cai_CompACT_ICCV_2015}, DeepParts \cite{Tian_DeepParts_ICCV_2015}, Checkerboards \cite{Zhang_FCF_CVPR_2015}, CCF+CF \cite{Yang_CCF_ICCV_2015}, LDCF \cite{Nam_LDCF_NIPS_2014}, and TA-CNN \cite{Tian_Ta_CVPR_2015}) on Caltech test dataset \cite{Dollar_PD_PAMI_2012}. In \cite{Dollar_PD_PAMI_2012}, it states that it is important
to detect pedestrians of near scale (pedestrians over 80 pixels) and medium scales (pedestrians over 30 pixels
and under 80 pixels). Thus,  a harder test set (pedestrians over 30 pixels) is used to evaluate detection performance, which corresponds to the union set of near scale set and medium scale set. To compare with state-of-the-art methods and achieve best detection performance, the images are twice upsampled in the training and test stages. Fig. \ref{fig06} shows the ROC of these methods. It can be seen that MHN-D stably outperforms all the other state-of-the-art methods. $MR$ of MHN-D is 27.96\% and that of F-DNN+SS \cite{Du_FDNN_WACV_2017} is 31.43\%. It means that MHN-D outperforms F-DNN+SS by 3.47\%. We also trains MHN-D on reasonable test set. $MR$ of MHN-D is 9.20\%, which also is 0.80\% lower than that of MSCNN \cite{Cai_MSCNN_ECCV_2016}.

\begin{table}
\renewcommand{\arraystretch}{1.3}
\caption{Detection results on Citypersons test set.}
\begin{center}
\footnotesize
\begin{tabular}{p{2.0cm}|c|c|c}
\hline
Name                           & Repulsion \cite{Wang_Repulsion_CVPR_2018}    & AdaptedRCNN \cite{Zhang_CityPersions_CVPR_2017}       & MHN-D \\
\hline
backbone                       &  ResNet50         & VGG16                  & VGG16\\
proposal number                &  --              & 2000                   & 200\\
scale                          & $\times$1.5       &  $\times$1.3           & $\times$1.3\\
\hline
Reasonable                     & 11.48             &  12.97                 & 12.92\\
Reasonable{\_}{small}          & 15.67             &  37.24                 & 17.24\\
Reasonable{\_}{occ}            & 52.59             &  50.47                 & \textbf{46.72}\\
Reasonable{\_}{all}            & 39.17             &  43.86                 & \textbf{39.16}\\
\hline
\end{tabular}
\end{center}
\label{tab08}
\end{table}

\subsection{Comparison to some state-of-the-art methods on Citypersons dataset}
In this subsection, the proposed method is further compared to some state-of-art methods (i.e., Repulsion Loss \cite{Wang_Repulsion_CVPR_2018} and AdaptedRCNN \cite{Zhang_CityPersions_CVPR_2017}) on Citypersons test set. In the training and test stage, the images are 1.3 times upsampled. Table \ref{tab08} shows the results of these methods. According to \cite{Zhang_CityPersions_CVPR_2017}, the different test subsets (i.e., \texttt{Reasonable}, \texttt{Reasonable{\_}small}, \texttt{Reasonable{\_}occ}, \texttt{all}) are used for detailed comparison. \texttt{Reasonable} test subset contains 50 pixels or taller, unoccluded or partial unoccluded pedestrians. \texttt{Reasonable{\_}small} test subset contains pedestrians over 50 pixels under 75 pixels. \texttt{Reasonable{\_}occ} test subset contains 50 pixels or taller, heavily unoccluded pedestrians. \texttt{All} test subset contains pedestrians over 20 pixels. Based on the same backbone (i.e., VGG16),  MHN-D almost outperforms adaptedRCNN on all four subsets. Specifically, MHN-D outperforms AdaptedRCNN by 0.05\%, 20.00\%, 3.75\%, and 4.70\% on \texttt{reasonable}, \texttt{reasonable{\_}small}, \texttt{reasonable{\_}occ}, and \texttt{all} test subsets, respectively.  Moreover, MHN-D only needs 200 proposals, while adaptedRCNN needs 2000 proposals. MHN-D has comparable detection performance with Repulsion \cite{Wang_Repulsion_CVPR_2018} which uses better backbone (i.e., ResNet50), a larger upsampled scale (i.e., $\times$1.5), and a better loss function.

\begin{table}
\renewcommand{\arraystretch}{1.25}
\caption{Detection performance of proposal generation and objection detection on COCO \texttt{minival} set. Average recall ($AR$) and average precision ($AP$) are used for evaluation. }
\begin{center}
\footnotesize
\begin{tabular}{p{1.3cm}|c|c|ccc}
\hline
\multicolumn{5}{l}{Proposal Generation}      \\
\hline
Name                         & backbone   & $AR$ & $AR_{s}$ & $AR_{m}$ & $AR_l$ \\
\hline
FPN \cite{Lin_FPN_CVPR_2017} & ResNet50   & 49.48   & 34.55& 55.70& 66.64 \\
MHN                          & ResNet50   & 50.76   & 37.19& 56.32& 66.76 \\
$\Delta$                     &   N/A        & \textbf{1.28}    & \textbf{2.64} & \textbf{0.62} &  0.12\\
\hline
\hline
\multicolumn{5}{l}{Object Detection}      \\
\hline
Name                         & backbone   & $AP$ & $AP_{s}$ & $AP_{m}$ & $AP_{l}$ \\
\hline
FPN \cite{Lin_FPN_CVPR_2017} & ResNet50   & 36.90   & 21.42& 40.01& 47.76 \\
MHN                          & ResNet50   & 38.20   & 23.49& 41.04& 47.81 \\
$\Delta$                     &    N/A   & \textbf{1.30}    & \textbf{2.07}& \textbf{1.03}& 0.05\\
\hline
FPN \cite{Lin_FPN_CVPR_2017} & ResNet101   & 39.04   & 22.63& 42.45& 50.66 \\
MHN                          & ResNet101   & 40.83   & 26.01& 44.46& 51.57 \\
$\Delta$                     &    N/A   & \textbf{1.79}    & \textbf{3.38}& \textbf{2.01}& 0.91\\
\hline
\end{tabular}
\end{center}
\label{tab09}
\end{table}

\begin{table*}
\renewcommand{\arraystretch}{1.25}
\caption{Single-model detection results of some state-of-the-art methods on COCO \texttt{test-dev} set.}
\begin{center}
\footnotesize
\begin{tabular}{p{3.0cm}|p{3.0cm}<{\centering}|ccc|ccc}
\hline
Method       & backbone                         & $AP$ & $AP_{50}$ & $AP_{75}$ & $AP_{s}$ & $AP_{m}$ & $AP_{l}$  \\
\hline
Faster RCNN \cite{He_ResNet_CVPR_2016}          & ResNet101 \cite{He_ResNet_CVPR_2016}          & 34.9 & 55.7 & 37.4 & 15.6 & 38.7 & 50.9\\
G-RMI \cite{Huang_SpeedAcc_CVPR_2017}           & Inception-ResNet-v2 \cite{Szegedy_G-RMI_AAAI_2017} & 34.7 & 55.5 & 36.7 & 13.5 & 38.1 & 52.0\\
TDM \cite{Shrivastava_TDM_arXiv_2016}           & Inception-ResNet-v2 \cite{Szegedy_G-RMI_AAAI_2017} & 36.8 & 57.7 & 39.2 & 16.2 & 39.8 & 52.1\\
FPN \cite{Lin_FPN_CVPR_2017}                    & ResNet101 \cite{He_ResNet_CVPR_2016}           & 36.2 & 59.1 & 39.0 & 18.2 & 39.0 & 48.2\\
RetinaNet \cite{Lin_Focal_ICCV_2017}            & ResNet101 \cite{He_ResNet_CVPR_2016}           & 37.8 & 57.5 & 40.8 & 20.2 & 41.1 & 49.2\\
RetinaNet \textit{ms-train} \cite{Lin_Focal_ICCV_2017}   & ResNet101 \cite{He_ResNet_CVPR_2016}           & 39.1 & 59.1 & 42.3 & 21.8 & 42.7 & 50.2\\
Mask RCNN \cite{He_MaskRCNN_ICCV_2017}          & ResNet101 \cite{He_ResNet_CVPR_2016}           & 38.2 & 60.3 & 41.7 & 20.1 & 41.1 & 50.2\\
Mask RCNN \cite{He_MaskRCNN_ICCV_2017}          & ResNeXt101 \cite{Xie_ResNeXt_CVPR_2017}         & 39.8 & 62.3 & 43.4 & 22.1 & 43.2 & 51.2\\
\hline
MHN           & ResNet50 \cite{He_ResNet_CVPR_2016}             & 38.7 & 61.3 & 41.9 & \textbf{23.9} & 41.0 & 47.1\\
MHN           & ResNet101 \cite{He_ResNet_CVPR_2016}            & \textbf{41.1} & \textbf{63.6} & \textbf{44.8} & \textbf{25.5} & \textbf{43.5} & 50.4\\
\hline
MHN \textit{ms-test}  & ResNet50 \cite{He_ResNet_CVPR_2016}     & 41.8 & 63.1 & 46.0 & 27.0 & 43.8 & 51.1\\
MHN \textit{ms-test}  & ResNet101 \cite{He_ResNet_CVPR_2016}    & 44.1 & 65.4 & 48.7 & 29.0 & 46.4 & 54.0\\
\hline
\end{tabular}
\end{center}
\label{tab10}
\end{table*}

\subsection{Experiments on COCO benchmark}
\label{ExpCOCO}
To further demonstrate suitability of proposed method for general object detection, the famous and challenging object detection dataset (i.e., COCO benchmark \cite{Lin_COCO_arXiv_2014}) is used. Based on detectron  platform \cite{Girshick_Detectron_2018}, FPN \cite{Lin_FPN_CVPR_2017} and MHN are reimplemented based on ResNet and Faster RCNN architecture \cite{He_ResNet_CVPR_2016} with similar parameter settings on two GPUs. The training and test images are rescaled so that the shorter side has 800 pixels. The number of total iterations is 360k, where the learning rate of first 240k iterations is 0.005, the learning rate of middle 80k iterations is 0.0005, and the learning rate of last 40k iterations is 0.00005. In each mini-batch, there are two images per GPU and 512 ROIs per image. FPN outputs proposals by five feature maps (i.e., P3-P7), and MHN outputs proposals by five feature maps (i.e., M3-M7). Table \ref{tab09} shows proposal generation results and object detection results of FPN and MHN on \texttt{minival} set. The results on small-scale objects, medium-scale objects, and large-scale objects are further given. It can be seen as follows: (1) MHN outperforms FPN on both proposal generation and object detection. Based on ResNet50, MHN outperforms FPN by 1.28\% and 1.30\% on proposal generation and object detection, respectively. (2) On small-scale object detection, MHN has much better detection performance. Based on ResNet50, MHN outperforms FPN by 2.64\% on proposal generation and 2.07\% on object detection. Based on ResNet101, MHN outperforms FPN by 3.38\% on object detection.

Table \ref{tab10} further compares MHN with some single model detection results of some state-of-the-art methods (i.e., Faster RCNN \cite{He_ResNet_CVPR_2016}, G-RMI \cite{Huang_SpeedAcc_CVPR_2017}, TDM \cite{Shrivastava_TDM_arXiv_2016}, FPN \cite{Lin_FPN_CVPR_2017}, RetinaNet \cite{Lin_Focal_ICCV_2017}, and Mask RCNN \cite{He_MaskRCNN_ICCV_2017}) on \texttt{test-dev} set. Among the these methods, our method achieves state-of-the-art performance. Based on the same model (i.e., ResNet101), MHN outperforms Mask RCNN by 2.9\%. Moreover, MHN has remarkably superior superior detection performance on small-scale object detection. For example, MHN outperforms Mask RCNN by 5.4\% on small-scale object detection. 

\section{Conclusion}
\label{Conclusion}
In this paper, we have proposed multi-branch and high-level semantic convolutional neural networks (MHN) for pedestrian detection. MHN builds multiple different branches, where the output maps of each branch have the same depth and high-level semantic features. Meanwhile, MHN uses skip-layer connections to add context information to the high-resolution branch for small-scale pedestrian detection. To take the advantage of dilated convolution, MHN-D incorporates it into part branch of MHN. When MHN and MHN-D are embedded into Faster RCNN architecture, the weighted scores of proposal generation and proposal classification are used for improving detection performance. Experiments on KITTI dataset \cite{Geiger_KITTI_CVPR_2012}, Caltech pedestrian dataset \cite{Dollar_PD_PAMI_2012}, and Citypersons dataset have demonstrated that the proposed methods are superior to other feature pyramid architectures (i.e., MSCNN \cite{Cai_MSCNN_ECCV_2016} and FPN \cite{Lin_FPN_CVPR_2017}) for pedestrian detection. Moreover, the proposed method is also useful on general object detection based on experiments on COCO benchmark.

\ifCLASSOPTIONcaptionsoff
  \newpage
\fi

\end{document}